\documentclass[a4paper,11pt]{article}
\usepackage[margin=2cm]{geometry}
\usepackage{amsfonts}
\usepackage{amsmath}
\usepackage{mathptmx}
\usepackage{amssymb}
\usepackage{graphics}
\usepackage{graphicx}
\usepackage{times}
\usepackage{listings}
\usepackage[linktocpage,colorlinks=true,urlcolor=blue,citecolor=red,bookmarks=false]{hyperref}
\usepackage[font=small,bf]{caption}

\usepackage{color}

\definecolor{mygreen}{rgb}{0,0.6,0}
\definecolor{mygray}{rgb}{0.5,0.5,0.5}
\definecolor{mymauve}{rgb}{0.58,0,0.82}

\title{gSLICr: SLIC superpixels at over 250Hz}
\author{
	Carl Yuheng Ren  \\ \url{carl@robots.ox.ac.uk}  \\ University of Oxford \and
	Victor Adrian Prisacariu \\ \url{victor@robots.ox.ac.uk} \\ University of Oxford \and
	Ian D Reid \\ \url{ian.reid@adelaide.edu.au} \\ University of Adelaide
}

\date{\today}

\begin{document}
\maketitle

\begin{abstract}
We introduce a parallel GPU implementation of the Simple Linear Iterative Clustering (SLIC) superpixel segmentation. Using a single graphic card, our implementation achieves speedups of up to $83\times$ from the standard sequential implementation. Our implementation is fully compatible with the standard sequential implementation and the software is now available online and is open source.
\end{abstract}

\tableofcontents

\section{Introduction}
\label{sec:intro}
Superpixels are regions of pixels grouped in some perceptually meaningful way, usually following colour or boundary cues. They are designed to produce a simpler and more compact representation for an image, while keeping its semantic meaning intact. Superpixel segmentation is used most often as an image preprocessing step, with a view towards reducing computational complexity for subsequent processing steps.

The term \textit{superpixel}, along with the first notable superpixel algorithm, was introduced by \cite{shi_tpami_2000}. Many algorithms followed, using various types of image features, various optimisation strategies and various implementations techniques. These algorithms have varying specifications and performance requirements. For example, some algorithms aim to find a fixed number of superpixels, others try to find the minimum possible number of superpixels by imposing a colour cohesion requirement, while others place emphasis on matching image boundaries. Sometimes fast processing is required, when for example, the superpixel algorithm is used as a precursor to a tracker. Sometimes superpixels are designed not to under-segment the image, when used for example as a means of condensing the image information, to serve as the basis of a labelling problem. Regardless of its design, superpixel segmentation is usually among the first steps in a much longer processing pipeline. Therefore, we believe that, for any superpixel method to be useful, it must satisfy the two following requirements:
\begin{itemize}
	\item it should not decrease the performance of the full processing pipeline;
	\item it should be fast.
\end{itemize}

The performance requirement is satisfied in many vision applications by superpixels that are \textit{compact}, \textit{uniform} and \textit{follow image edges}. These requirements motivated the \textbf{simple iterative clustering algorithm} (SLIC) algorithm of \cite{achanta_tpami_2012}. This is simple, efficient and suitable for real-time operation. Still however, the CPU-sequential implementation of SLIC need 300$\sim$400ms to segment a single 640x480 image. Reducing the number of iterations for each clustering pass can make the algorithm faster, at the cost of a decrease in performance.

In this work, we propose a GPU implementation of the SLIC algorithm, using the NVIDIA CUDA framework. Our implementation is up to $83\times$ faster than the original CPU implementation of \cite{achanta_tpami_2012}, making it, to our knowledge, the fastest superpixel method to date.

Our full source code with a simple example can be downloaded from \url{https://github.com/carlren/gSLICr}. The following sections describe in detail our algorithm and implementation.

\section{Simple Linear Iterative Clustering (SLIC)}
\label{sec:SLIC}

\begin{figure}
\centering
\begin{tabular}{cc}
  \fbox{\includegraphics[width=0.45\textwidth]{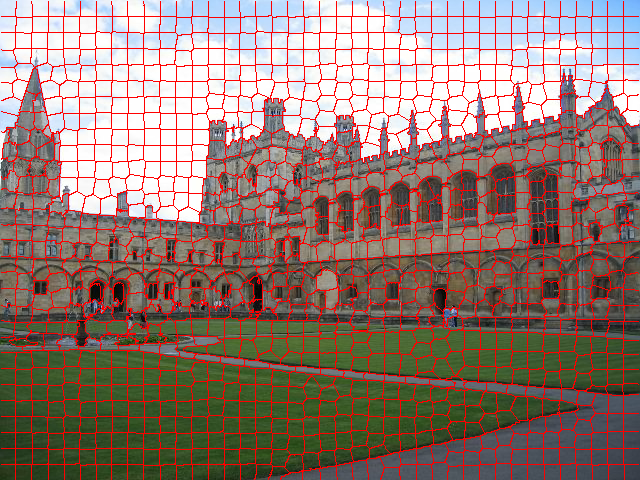}} &
  \fbox{\includegraphics[width=0.45\textwidth]{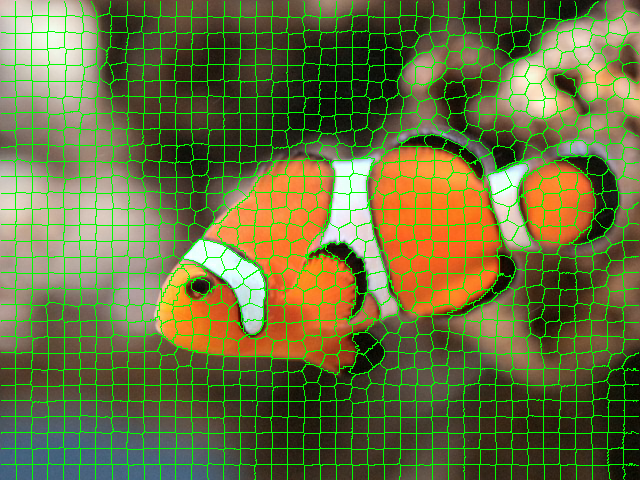}}
\end{tabular}
\caption{Example SLIC superpixel segmentation}
\label{fig:SLIC_sample}
\end{figure}

The Simple Linear Iterative Clustering (SLIC) algorithm for superpixel segmentation was proposed in \cite{achanta_tpami_2012}. An example segmentation result is shown in \ref{fig:SLIC_sample}.

SLIC uses a k-means-based approach to build local clusters of pixels in the 5D $[labxy]$ space defined by the $L,a,b$ values of the CIELAB color space and the $x,y$ pixel coordinates. The CIELAB color space is chosen because it is perceptually uniform for a small distance in colour. 

SLIC uses as input the desired number of approximately equally-sized superpixels $K$. Given an image with $N$ pixels, the approximate size of each superpixel therefore is $N/K$. Assuming roughly equally sized superpixels, there would be a superpixel center at every grid interval $S=\sqrt{N/K}$. Let $[l_i,a_i,b_i,x_i,y_i]^T$ be the 5D point corresponding to a pixel. Writing the cluster center $C_k$ as $C_k = [l_k,a_k,b_k,x_k,y_k]^T$, SLIC defines a distance measure $D_k$ as:
\begin{eqnarray}\label{equ:Ds_Define}
        d_{lab}=\sqrt{(l_k-l_i)^2+(a_k-a_i)^2+(n_k-b_i)^2}\nonumber\\
        d_{xy}=\sqrt{(x_k-x_i)^2+(y_k-y_i)^2}\nonumber\\
        D_s=d_{lab}+\frac{m}{S}d_{xy}
\end{eqnarray}
where $D_s$ is the sum of the $lab$ distance and the $xy$ plane distance \textit{normalized} by the grid interval $S$. The variable $m$ controls the compactness of superpixels i.e. the greater the value of $m$, the more spatial proximity is emphasized and thus the more compact the cluster becomes.

This distance metric is next used in a standard local k-means algorithm. First, the cluster centers are perturbed to the lowers gradient position from a local neighborhood. Next, iteratively, the algorithm assigns the best matching pixels in a local neighborhood to each cluster and computes new center locations. The process is stopped when the cluster centers stabilise i.e. when the L1 distance between centers at consecutive iterations is smaller than a set threshold. Finally, connectively is enforced to cleanup the final superpixel lattice.

In the next section we detail our GPU implementation of SLIC.

\section{gSLICr GPU Implementation}
\label{sec:gSLIC_impl}

\begin{figure}
	\centering
	\begin{tabular}{cc}
		\includegraphics[width=0.6\textwidth]{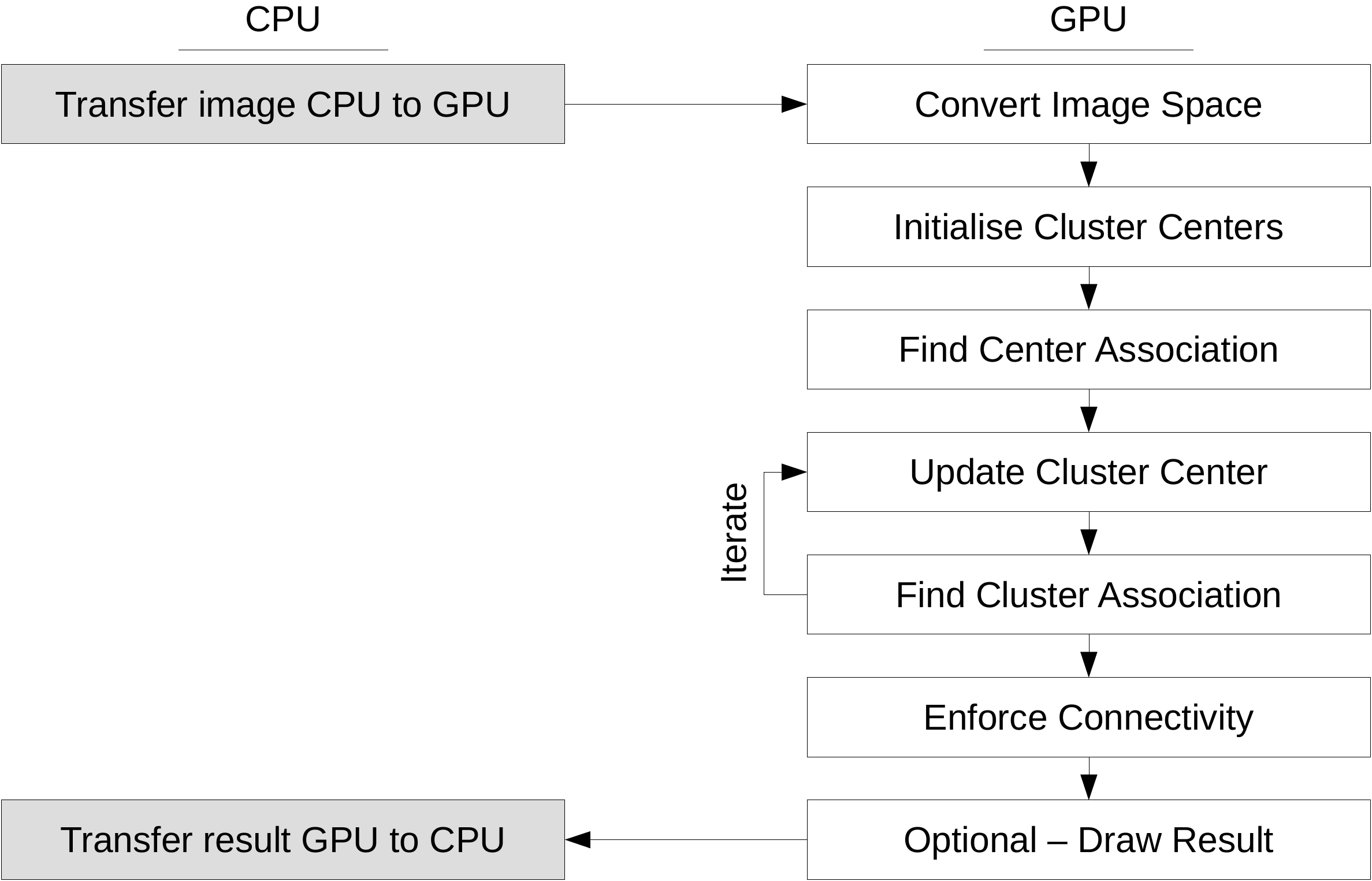}
	\end{tabular}
	\caption{Workflow of gSLICr}
	\label{fig:gSLICr_flowchart}
\end{figure}

We split our implementation into two parts, as shown in Figure \ref{fig:gSLICr_flowchart}: the GPU is responsible for most of the processing, with only data acquisition and visualization being left for the CPU.

The GPU implementation then proceeds as follows:

\begin{itemize}
\item \textbf{Image space conversion}: The RGB input image is converted to Cielab, using one thread for each pixel.

\item \textbf{Cluster center initialisation}: We use one thread per cluster center (i.e. superpixel) to initialise our superpixel map. This is an $ns_r \times ns_c$ image which contains, for each entry, center coordinates, number of associated pixels and colour information. $ns_r$ and $ns_c$ represent the number of superpixels per image row and column, respectively. 

\item \textbf{Finding the cluster associations}: Each pixel in the image determines what is its closest cluster using the 5D distance detailed in the previous section. This requires a maximum of nine cluster centers to be examined and is done using one thread per pixel.

\item \textbf{Updating the cluster center}: Here we update each cluster center using the pixels assigned to it. This process is done in two separate kernels. First, each cluster center must access all pixels associated to it, within a local neighborhood that is a function of the superpixel size. Here we use $ns_r \times ns_c \times n_{bl}$, where $ns_r$ and $ns_c$ are defined as before. $n_{bl} = \text{spixel\_size} \times 3 / \text{BLOCK\_DIM}$ captures the number of pixels each thread can process, as a function of superpixel size and thread block dimension (16 in our case). The result is written to an image of size $ns_r \times ns_c \times n_{bl}$, upon which we run a reduction step on the third dimension to obtain the final updated cluster center positions.

\item \textbf{Enforce connectivity}: We eliminate stray pixels with two one thread per pixel calls of the same kernel. This prompts a pixel to change its label of that of the surrounding pixels (in a $2\times2$ neighborhood) if all have a different label.
\end{itemize}

\section{Library Usage}
\label{sec:lib_useage}
Our full code can be downloaded from \url{https://github.com/carlren/gSLICr}. It consists of (i) a demo project (which requires OpenCV) and (ii) a separate library (which has no dependencies). 

The demo project acquires images from the camera, processes them through the library and displays the result back in an OpenCV window. It creates an instances of the \textbf{core\_engine} class, which is the main access point to our code. This controls the segmentation code in the \textbf{seg\_engine} class and times the result. 

The \textbf{seg\_engine} class is responsible for all the superpixel processing, and the algorithm is controlled from the \textbf{Perform\_Segmentation} method. This code is listed below:

\begin{lstlisting}
void seg_engine::Perform_Segmentation(UChar4Image* in_img)
{
	source_img->SetFrom(in_img, ORUtils::MemoryBlock<Vector4u>::CPU_TO_CUDA);
	Cvt_Img_Space(source_img, cvt_img, gSLICr_settings.color_space);

	Init_Cluster_Centers();
	Find_Center_Association();

	for (int i = 0; i < gSLICr_settings.no_iters; i++)
	{
		Update_Cluster_Center();
		Find_Center_Association();
	}

	if(gSLICr_settings.do_enforce_connectivity) Enforce_Connectivity();
	cudaThreadSynchronize();
}
\end{lstlisting}

\begin{figure}
	\centering
	\begin{tabular}{cc}
		\includegraphics[width=0.6\textwidth]{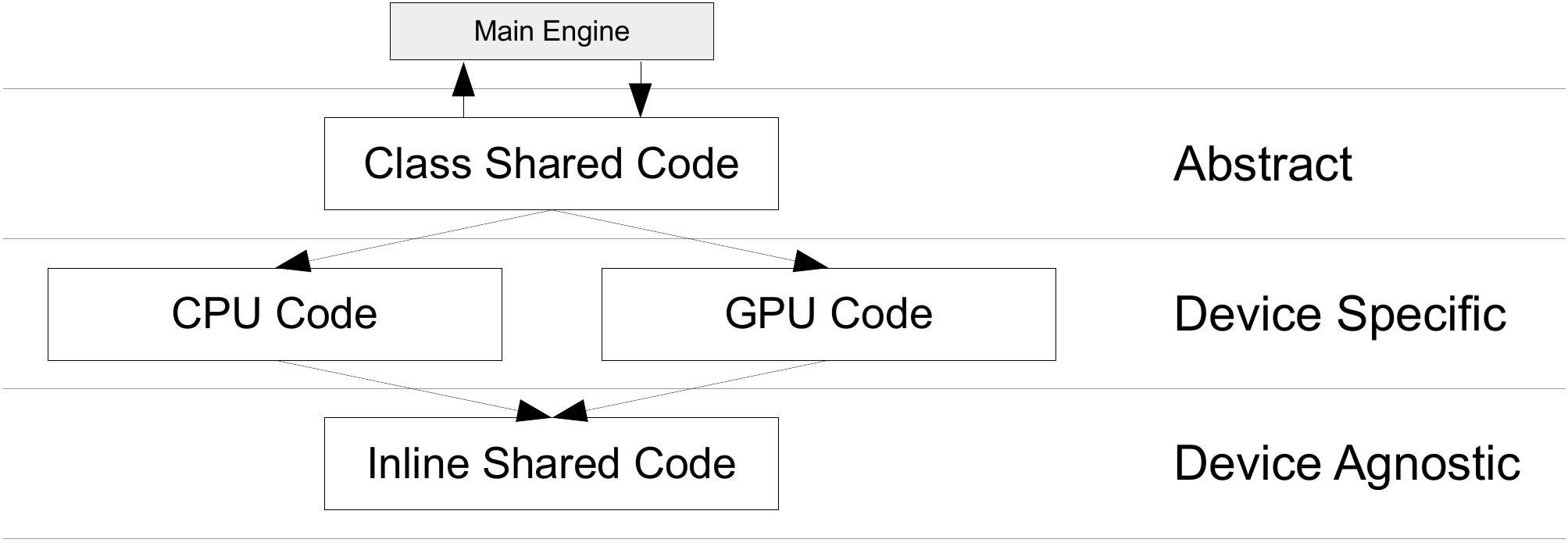}
	\end{tabular}
	\caption{Cross Device Engine Design Pattern}
	\label{fig:design_pattern}
\end{figure}

Similar to the other projects from the Oxford Vision Library, such as LibISR \cite{star3d_iccv_2013} or InfiniTAM \cite{PrisacariuKCRVTRM14}, this class follows our cross device engine design pattern outlined in Figure \ref{fig:design_pattern}. The engine is split into 3 layers. The topmost, so called Abstract Layer, contains the main algorithm function calls (listed above). The abstract interface is implemented in the next, Device Specific Layer, which may be very different between e.g. a CPU and a GPU implementation. Further implementations using e.g. OpenMP or other hardware acceleration architectures are possible. We only provide a GPU implementation in this case, in the \textbf{gSLICr\_seg\_engine\_GPU.h} and \textbf{gSLICr\_seg\_engine\_GPU.cu} files. At the third, Device Agnostic Layer, there is some inline C-code that may be called from the higher layers. This contains the bulk of the per-pixel and per-cluster processing code and can be found in the \textbf{gSLICr\_seg\_engine\_shared.h} file.

\section{Results}
\label{sec:results}

Our implementation is designed to produce the same result as the sequential SLIC implementation of \cite{achanta_tpami_2012}, so qualitatively and quantitatively the results are virtually identical. Our method is however considerably faster than this, and, to our knowledge, all other superpixel segmentation techniques. We included a comparison with many such techniques in Table \ref{tab:timings}. Here we used four images sizes, and our method was consistently much faster than any other approach. Compared to the original SLIC algorithm, our approach is up to $83\times$ faster. 

\section*{Acknowldgements}
We thank Magellium ltd. and CNAS for their support for our experimental section.

\newpage

\begin{table}[htpb]
	\centering
	\begin{tabular}{|c|ccccc|}
		\hline method & $1024\times1024$ & $3631\times3859$ & $963\times1024$ & $1002\times1002$ & $933\times800$ \\ \hline
		\hline \parbox[c][0.95cm]{2.50cm}{\centering \textbf{Current Work} \\ 1000 spx} & 0.01 & 0.12 & 0.01 & 0.01 & 0.008 \\
		\hline \parbox[c][0.95cm]{2.50cm}{\centering \textbf{Current Work} \\ 2000 spx} & 0.01 & 0.12 & 0.01 & 0.01 & 0.008 \\
		\hline \parbox[c][0.95cm]{2.50cm}{\centering Achanta \cite{achanta_tpami_2012} \\ 1000 spx} & 0.73 & 9.69 & 0.75 & 0.70 & 0.55 \\
		\hline \parbox[c][0.95cm]{2.50cm}{\centering Achanta \cite{achanta_tpami_2012} \\ 2000 spx} & 0.74 & 9.88 & 0.72 & 0.71 & 0.55 \\
		\hline \parbox[c][0.95cm]{2.50cm}{\centering Veksler \cite{veksler_eccv_2010} \\ patch size 25} & 22.6 & 321 & 20.6 & 19.3 & 16.7 \\
		\hline \parbox[c][0.95cm]{2.50cm}{\centering Veksler \cite{veksler_eccv_2010} \\ patch size 50} & 25.2 & 368 & 21.6 & 19.7 & 17.7 \\
		\hline \parbox[c][0.95cm]{2.50cm}{\centering Zhang \cite{zhang_iccv_2011} \\ patch size 40} & 0.63 & 7.49 & 0.49 & 0.50 & 0.36 \\
		\hline \parbox[c][0.95cm]{2.50cm}{\centering Zhang \cite{zhang_iccv_2011} \\ patch size 60} & 0.60 & 7.53 & 0.48 & 0.50 & 0.36 \\
		\hline \parbox[c][0.95cm]{2.50cm}{\centering Comaniciu \cite{comaniciu_tpami_2002} \\ sp. bandw. 11} & 12.1 & x & 12.3 & 9.73 & 9.47 \\
		\hline \parbox[c][0.95cm]{2.50cm}{\centering Comaniciu \cite{comaniciu_tpami_2002} \\ sp. bandw. 23} & 43.3 & x & 53.8 & 32.5 & 34.2 \\
		\hline \parbox[c][0.95cm]{2.50cm}{\centering Shi \cite{shi_tpami_2000} \\ 1000 spx} & 329 & x & 316 & 334 & 218 \\
		\hline \parbox[c][0.95cm]{2.50cm}{\centering Shi \cite{shi_tpami_2000} \\ 2000 spx} & 515 & x & 384 & 411 & 338 \\
		\hline \parbox[c][0.95cm]{2.50cm}{\centering Felzenszwalb \\ \cite{felzenszwalb_ijcv_2004} $\sigma = 0.4$} & 0.62 & 17.4 & 0.96 & 0.85 & 0.41 \\
		\hline \parbox[c][0.95cm]{2.50cm}{\centering Felzenszwalb \\ \cite{felzenszwalb_ijcv_2004} $\sigma = 0.5$} & 0.70 & 18.3 & 0.97 & 0.92 & 0.47 \\
		\hline \parbox[c][0.95cm]{2.50cm}{\centering Liu \cite{lin_cvpr_2011} \\ 500 spx} & 6.76 & 116 & 7.34 & 7.00 & 5.34 \\
		\hline \parbox[c][0.95cm]{2.50cm}{\centering Liu \cite{lin_cvpr_2011} \\ 1000 spx} & 6.94 & 116 & 6.90 & 7.10 & 5.35 \\
		\hline \parbox[c][0.95cm]{2.50cm}{\centering Liu \cite{lin_cvpr_2011} \\ 2000 spx} & 7.26 & 120 & 7.23 & 7.60 & 5.62 \\
		\hline \parbox[c][0.95cm]{2.50cm}{\centering den Bergh \cite{bergh_eccv_2012} \\ 200 spx} & 1.53 & 22.1 & 1.67 & 1.39 & 1.01 \\
		\hline \parbox[c][0.95cm]{2.50cm}{\centering den Bergh \cite{bergh_eccv_2012} \\ 400 spx} & 2.36 & 31.2 & 2.35 & 2.00 & 1.42 \\
		\hline \parbox[c][0.95cm]{2.50cm}{\centering Levinshtein \\ \cite{levinshtein_tpami_2009} 1000 spx} & 38.7 & x & 46.6 & 52.1 & 26.9 \\
		\hline \parbox[c][0.95cm]{2.50cm}{\centering Levinshtein \\ \cite{levinshtein_tpami_2009} 2000 spx} & 38.4 & x & 44.0 & 55.6 & 26.8 \\
		\hline \parbox[c][0.95cm]{2.50cm}{\centering Moore \cite{moore_cvpr_2008} \\ bounds \cite{leordean_eccv_2012}} & 2.07 & x & 2.20 & 1.79 & 1.37 \\
		\hline \parbox[c][0.95cm]{2.50cm}{\centering Moore \cite{moore_cvpr_2008} \\ bounds \cite{dollar_cvpr_2006}} & 2.44 & x & 2.12 & 2.17 & 1.25 \\
		\hline
	\end{tabular}
	\caption{Timing results for the tested methods.}
	\label{tab:timings}
\end{table}

\bibliographystyle{}

\begin{thebibliography}{10}
	
	\bibitem{shi_tpami_2000}
	J.~Shi and J.~Malik, ``{Normalized Cuts and Image Segmentation},'' {\em
		T-PAMI}, vol.~22, pp.~888--905, 2000.
	
	\bibitem{achanta_tpami_2012}
	R.~Achanta, A.~Shaji, K.~Smith, A.~Lucchi, P.~Fua, and S.~S{\"u}sstrunk,
	``{SLIC Superpixels Compared to State-of-the-Art Superpixel Methods},'' {\em
		T-PAMI}, vol.~34, no.~11, pp.~2274--2282, 2012.
	
	\bibitem{star3d_iccv_2013}
	C.~Ren, V.~Prisacariu, D.~Murray, and I.~Reid, ``Star3d: Simultaneous tracking
	and reconstruction of 3d objects using rgb-d data,'' in {\em Computer Vision
		(ICCV), 2013 IEEE International Conference on}, pp.~1561--1568, Dec 2013.
	
	\bibitem{PrisacariuKCRVTRM14}
	V.~A. Prisacariu, O.~K{\"{a}}hler, M.~Cheng, C.~Y. Ren, J.~P.~C. Valentin,
	P.~H.~S. Torr, I.~D. Reid, and D.~W. Murray, ``A framework for the volumetric
	integration of depth images,'' {\em CoRR}, vol.~abs/1410.0925, 2014.
	
	\bibitem{veksler_eccv_2010}
	O.~Veksler, Y.~Boykov, and P.~Mehrani in {\em {ECCV}}, pp.~211--224, 2010.
	
	\bibitem{zhang_iccv_2011}
	Y.~Zhang, R.~I. Hartley, J.~Mashford, and S.~Burn, ``{Superpixels via
		pseudo-Boolean optimization},'' in {\em {ICCV}}, pp.~1387--1394, 2011.
	
	\bibitem{comaniciu_tpami_2002}
	D.~Comaniciu and P.~Meer, ``{Mean Shift: A Robust Approach Toward Feature Space
		Analysis},'' {\em T-PAMI}, vol.~24, no.~5, pp.~603--619, 2002.
	
	\bibitem{felzenszwalb_ijcv_2004}
	P.~F. Felzenszwalb and D.~P. Huttenlocher, ``{Efficient Graph-Based Image
		Segmentation},'' {\em IJCV}, vol.~59, no.~2, pp.~167--181, 2004.
	
	\bibitem{lin_cvpr_2011}
	M.-Y. Liu, O.~Tuzel, S.~Ramalingam, and R.~Chellappa, ``{Entropy rate
		superpixel segmentation},'' in {\em {CVPR}}, pp.~2097--2104, 2011.
	
	\bibitem{bergh_eccv_2012}
	M.~V. den Bergh, X.~Boix, G.~Roig, B.~de~Capitani, and L.~J.~V. Gool, ``{SEEDS:
		Superpixels Extracted via Energy-Driven Sampling},'' in {\em {ECCV}},
	pp.~13--26, 2012.
	
	\bibitem{levinshtein_tpami_2009}
	A.~Levinshtein, A.~Stere, K.~N. Kutulakos, D.~J. Fleet, S.~J. Dickinson, and
	K.~Siddiqi, ``{TurboPixels: Fast Superpixels Using Geometric Flows},'' {\em
		T-PAMI}, vol.~31, no.~12, pp.~2290--2297, 2009.
	
	\bibitem{moore_cvpr_2008}
	A.~P. Moore, S.~J.~D. Prince, J.~Warrell, U.~Mohammed, and G.~Jones,
	``{Superpixel lattices},'' in {\em CVPR}, 2008.
	
	\bibitem{leordean_eccv_2012}
	M.~Leordeanu, R.~Sukthankar, and C.~Sminchisescu, ``{Efficient closed-form
		solution to generalized boundary detection},'' in {\em ECCV}, pp.~516--529,
	2012.
	
	\bibitem{dollar_cvpr_2006}
	P.~Dollar, Z.~Tu, and S.~Belongie, ``{Supervised Learning of Edges and Object
		Boundaries},'' in {\em CVPR}, vol.~2, pp.~1964--1971, 2006.
	
\end{thebibliography}

\end{document}